# N-fold Superposition: Improving Neural Networks by Reducing the Noise in Feature Maps

Yang Liu, Qiang Qu, Chao Gao

National Digital Switching System Engineering & Technological R&D Center, Zhengzhou 450002, China

fabyangliu@hotmail.com

*Abstract*—Considering the use of Fully Connected (FC) layer limits the performance of Convolutional Neural Networks (CNNs), this paper develops a method to improve the coupling between the convolution layer and the FC layer by reducing the noise in Feature Maps (FMs). Our approach is divided into three steps. Firstly, we separate all the FMs into n blocks equally. Then, the weighted summation of FMs at the same position in all blocks constitutes a new block of FMs. Finally, we replicate this new block into n copies and concatenate them as the input to the FC layer. This sharing of FMs could reduce the noise in them apparently and avert the impact by a particular FM on the specific part weight of hidden layers, hence preventing the network from overfitting to some extent. Using the Fermat Lemma, we prove that this method could make the global minima value range of the loss function wider, by which makes it easier for neural networks to converge and accelerates the convergence process. This method does not significantly increase the amounts of network parameters (only a few more coefficients added), and the experiments demonstrate that this method could increase the convergence speed and improve the classification performance of neural networks.

## I. INTRODUCTION

Neural networks especially Convolutional Neural Networks (CNNs), have shown their remarkable performance in diverse domains of computer vision [1]. The reason for such impressive success CNNs have achieved mainly due to its specially designed network structure, e.g., convolution, pooling. Feature Maps (FMs), the abstract representation of the input image, could be obtained by the convolution operation. With the stack of convolution layers, high-level abstract feature representation can be secured to understand and identify the input image [2].

However, the use of coupling between the convolution layer and the fully connected (FC) layer is the main reason conventional CNNs overfits to the data or easily trapped in local minima, with poor predictions [3, 4]. Many methods have been developed in recent years to address these problems and improve the performance of CNNs. These methods mainly focus on the modifying of the network structure and regularization strategies.

Reference [4-6] replace the conventional convolution structure with a more vigorous approximation of a nonlinear function, which helps convolution layers capture a higher level of abstraction. However, this increases the amounts of network parameters and computational complexity. The pooling layer is usually used to abstract FMs to reduce overfitting, and the commonly used pooling methods are max pooling and average pooling [7]. Global average pooling [4] has been successfully used in most fairly known convolutional neural networks [8-10]. This method sums out the spatial information of each FM, therefore, reinforces similarities and meanwhile reduce differences in the spatial information, and makes it more robust to spatial translations of the input (which is very desirable). Reference [4] even tried to replace FC layer with this method to erase the effect of FC layer on the classification performance. But this makes it harder for neural networks to learn thus slow down the convergence process.

Softmax loss function is wildly used in most CNNs [11]; nevertheless, it is biased to the sample distribution, so it also becomes a major improvement goal for researchers. By adding a decision variable to softmax loss, the loss function could be explicitly encouraged intra-class compactness and inter-class separability between learned features, with avoiding overfitting [12]. However, this method needs repeated fine-tuning which makes the training difficult. Regularization term could also be added as a constraint to the loss function to prevent the model from overfitting, such as squared $L_2$ norm constraint on the weight [13]. This kind of regularization method intends to smaller the network weights and make the model simple to reduce overfitting. Dropout [14, 15] randomly drop units (along with their connections) from the network during training. Dropout forces the randomly selected neurons to work together to prevent units from co-adapting too much and improves the generalization ability of models.

Other strategies, e.g., batch normalization [16] reduces the internal-covariate-shift by normalizing input distributions of every layer to the standard Gaussian distribution. Initialization methods [17, 18] derive more robust initialization method that particularly considers the nonlinearities in neural networks. Data augmentation methods [19, 20] could make the model more robust and prevent overfitting when the training set is limited. These methods also provide us with valid attempts to improve the performance of neural networks, besides the modification of the network structure and regularization.

In a word, people have made various attempts to improve the performance of CNNs. In this paper, we propose a simple method named N-fold Superposition (NS) to enhance the performance of CNNs using the FC layer. By the weighted sum of multiple FMs, NS could reduce the noise in them, which generalizes the dependency between FMs and hidden units, and improves the coupling between the convolution layer and the FC layer, thus reducing overfitting to some degree. Our theoretical analysis proves that NS could make the model easier to converge and improve the performance of the network by constructing more global minimum points.

## II. Related Work

NS could also be interpreted as a way of regularizing the dependence between FMs and the weights in FC layer in neural networks. In this paper, we will mainly compare our approach with other regularization methods. The regularization methods primarily include $L_1$ norm [21], $L_2$ norm (weight decay) and Dropout, and are useful techniques for reducing generalization errors. Other approaches could also avoid overfitting, e.g., early stopping [22] and data augmentation (but are not the regularization methods we compare with).

The $L_1$ and $L_2$ norm methods add a regularization term to the loss function: the former adds the weighted sum of the absolute values of all the weights, and the latter adds the weighted sum of all the weights' square values. Then through backpropagation, the weight of the network will tend to 0. However, in reality, we hope that this process is not blind, the gradient should contain more information, and the process has stronger resistance to noise.

Reference [15] deems that standard backpropagation learning builds up brittle co-adaptations of hidden units which leads to overfitting, and the Dropout breaks up these co-adaptations by making the presence of any particular hidden unit unreliable. For the CNNs with the FC layer as their classification layer, the dependency between certain-part weights of hidden layers and a single FM will quickly lead up to brittle co-adaptations.

Local Response Normalization (LRN) [19] implements a form of lateral inhibition inspired by the type found in real neurons, which normalizes local responses within channels or between channels. However, this kind of normalization increases the training difficulty due to the need to fine-tune too many hyper-parameters, and the bioinspired interpretation is kindly far-fetched, and meanwhile, it increases the calculation complexity of backpropagation.

## III. Methodology

Considering the using of FC layers is the main reason for the overfitting problem of CNNs [4], and under the inspiration of the methods discussed previously, we developed the NS method. Fig.1 shows the operation process of NS. We simplify the network to a model with only one convolution layer and one FC layer.

Firstly, we can get FM $I^l$ by convolving the input $X$ with the convolution kernel $K^l$ using zero-padding, and the convolution stride is fixed to 1 pixel,

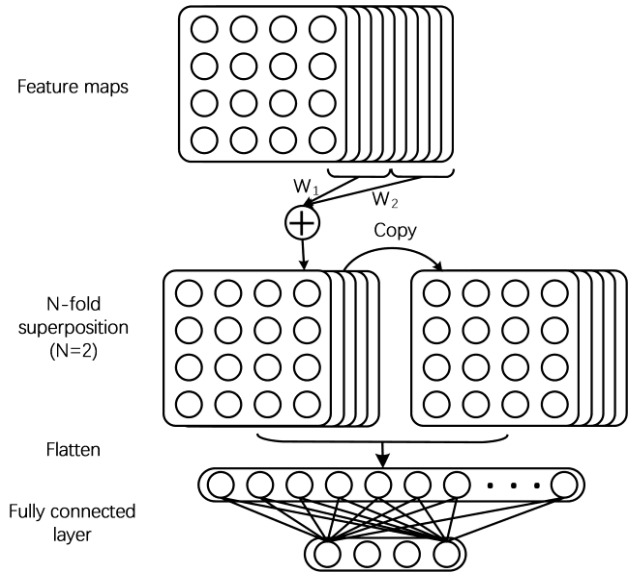

Fig. 1. The operation process of NS. This example excludes other operations like pooling, activation etc. to better illustrate the operation process of NS. In this paper, $N$ is set to 2 or 4.

$$I^l = g(X, K^l) \qquad (1)$$

where $X \in R^{v \times u}$, $K^l \in R^{m \times n}$, $I^l \in R^{(v-m+1) \times (u-n+1)}$, $l$ denotes the index of the convolution kernel and its corresponding FM ($1 \leq l \leq t$), $\{v, u, l, m, n\} \in N^*$. $g(\cdot)$ represents convolution function.

Secondly, NS divides all the FMs into $N$ blocks equally. The weighted summation of FMs at the same position of all the blocks constitutes new FMs at that position, and these new FMs form up a new block, as shown in (2),

$$\{I^1, ..., I^{\frac{t}{N}}\} \to \{\sum_{r=0}^{N-1} \beta_r I^{1+r*\frac{t}{N}}, ..., \sum_{r=0}^{N-1} \beta_r I^{(r+1)*\frac{t}{N}}\} \qquad (2)$$

where $\beta_r \in R$ ($r = 0, ..., N-1$) is the weighting coefficient of FMs, and $t$ have to be divisible by $N$ ($N \in N^*$).

Finally, we copy this new block $N$ times and concatenate them as the input to the FC layer,

$$C^{l+k*\frac{t}{N}} = f(\sum_{r=0}^{N-1} \beta_r I^{l+r*\frac{t}{N}}) \qquad (3)$$

where $C^{l+k*\frac{t}{N}} \in R^{1 \times (v-m+1) \cdot (u-n+1)}$ ($l = 1, ..., \frac{t}{N}; k = 0, ..., N-1$). And function $f(\cdot)$ flattens the input FM.

We introduce the NS method with two strategies including the Fixed-coefficient N-fold Superposition (FNS, all the weighted coefficient $\beta$ are set to the same value manually) and the Trainable-coefficient N-fold Superposition (TNS, all the weighted coefficient $\beta$ are set to be trainable).

## A. FM Sharing & Hidden Layer Weight Sharing

Assume that output has ten categories, the hidden layer weight $W \in R^{t \cdot (v-m+1) \cdot (u-n+1) \times 10}$, and $b \in R$ ($b$ is set as a uniform variable). The output of the FC layer is

$$y^o = \sum_{l=1}^{t} C^l W_{1+(l-1) \cdot w : l \cdot w, o} + b \quad (4)$$

where $w = (v-m+1) \cdot (u-n+1)$, $o = 1,...,10$.

Loss function is cross-entropy based on softmax

$$y_{out}^o = \frac{e^{y^o}}{\sum_{i=1}^{10} e^{y^i}} \quad (5)$$

and the loss is

$$L = -\sum_{o=1}^{10} y_{label}^o \ln(y_{out}^o) \quad (6)$$

Suppose that $y_{label} = [y_{label}^1,...,y_{label}^{10}] = [1,0,...,0]$, then the loss becomes

$$L = -\ln(y_{out}^1) = -y^1 + \ln(\sum_{i=1}^{10} e^{y^i}) \quad (7)$$

Therefore:

$$\nabla_{W_{1+(l-1) \cdot w : l \cdot w, o}} L = \begin{cases} -C^l + \dfrac{e^{y^1} C^l}{\sum_{i=1}^{10} e^{y^i}}, & o = 1 \\[2ex] \dfrac{e^{y^o} C^l}{\sum_{i=1}^{10} e^{y^i}}, & o \neq 1 \end{cases} \quad (l = 1,...,t) \quad (8)$$

By (2), (3) and (8), the gradient update of the weight (in the FC layer) is impacted by multiple FMs. Due to this sharing of FMs, the similarities are strengthened, and the differences are diminished, among these FMs. Thus, it will weaken the noise in the FMs.

Similarly,

$$\nabla_{K^{l+k*\frac{t}{N}}} L = \nabla_{K^{l+k*\frac{t}{N}}} C^{l+k*\frac{t}{N}} \sum_{r=0}^{N-1} (-W_{1+(l+r*\frac{t}{N}-1) \cdot w : (l+r*\frac{t}{N}) \cdot w, 1} + \frac{\sum_{i=1}^{10} e^{y^i} W_{1+(l+r*\frac{t}{N}-1) \cdot w : (l+r*\frac{t}{N}) \cdot w, i}}{\sum_{j=1}^{10} e^{y^j}}) \quad (9)$$

where $l = 1,...,\dfrac{t}{N}$, $C^{l+k*\frac{t}{N}}$ satisfies (3).

However, the original gradient on the convolutional kernel is

$$\nabla_{K^{l+k*\frac{t}{N}}} L' = \nabla_{K^{l+k*\frac{t}{N}}} C^{l+k*\frac{t}{N}} (-W_{1+(l+k*\frac{t}{N}-1) \cdot w : (l+k*\frac{t}{N}) \cdot w, 1} + \frac{\sum_{i=1}^{10} e^{y^i} W_{1+(l+k*\frac{t}{N}-1) \cdot w : (l+k*\frac{t}{N}) \cdot w, i}}{\sum_{j=1}^{10} e^{y^j}}) \quad (10)$$

where $l = 1,...,\dfrac{t}{N}$, $\nabla_{K^{l+k*\frac{t}{N}}} C^{l+k*\frac{t}{N}}$ is a constant and

$$C^{l+k*\frac{t}{N}} = f(I^{l+k*\frac{t}{N}}) (l = 1,...,\dfrac{t}{N}) \quad (11)$$

Additionally, because $\nabla_{K^{l+k*\frac{t}{N}}} C^{l+k*\frac{t}{N}}$ is only related to the input $X$, it is treated as a constant.

By (10) and (11), the original gradient dependence of the convolutional kernel is brittle and straightforward, because a considerable noise on specific weight (for the case in (10) as an example) has a much and direct effect on the corresponding kernel. While the hidden layer weights (in different locations) are shared for the gradients update of convolution kernels, see in (9). Therefore, hidden layer weight sharing will share the similarities with FM sharing in the reduction of the noise in hidden layer weights.

Through the sharing of FMs and hidden layer weights, the gradients would contain more information. Also, the coupling between the convolution layer and the FC layer has stronger resistance to noise (which is more like a voting system, and most votes resist the noise of a few votes). Meanwhile, by synthesizing multiple FMs, NS could extend and strengthen the dependency between FMs and hidden layer weights, thus making their co-adaptations more generalize.

## B. The Proof that NS Extends the Range of Loss Function's Global Minima Value

*Theorem 1 (Fermat lemma):* Suppose that $x_0$ is the extreme point of the function $f$ in $I$, and $x_0$ is the interior point of $I$. If $f$ is differential at $x_0$, $f'(x_0) = 0$.

*Theorem 2 (The solution number of essential solutions of homogeneous linear equations):* For the homogeneous linear equations $A_{m \times n} X = 0$, if $r(A_{m \times n}) = r < n$, it has an essential solution system and the dimension number (or the number of solutions) of the essential solutions is $n - r$ ($r$ denotes the rank of the coefficient matrix $A_{m \times n}$).

*Inference 1 (Necessary conditions for multivariate function extremum):* Suppose $g : A \to R$ is a multivariate function, and $A$ is the subset of $R^n$. $x^0 = (x_1^0,...,x_n^0)$ is an extreme point of $g$. If $x^0$ is the interior point of $A$ and $g$ has the first order partial derivative, then $g'_{x_1^0}(x^0) = ... = g'_{x_n^0}(x^0) = 0$.

*Proof:* We build a univariate function $\phi(x)$, where $x$ is the variate and $\phi(x) = g(x, x_2^0,...,x_n^0)$. Therefore, $x_1^0$ is the extreme point of function $\phi(x)$ (which is differential to $x$). By *Theorem 1*, we get $\phi'(x_1^0) = 0$, that is, $g'_{x_1^0}(x^0) = 0$. Similarly, we could get $g'_{x_2^0}(x^0) = ... = g'_{x_n^0}(x^0) = 0$.

*Proof (NS Extends the Range of Loss Function's Global Minima Value):*

Before the nuclear transformation, the model satisfies (1), (4), (5), (6) and (11). Let $y_{label} = [y_{label}^1,...,y_{label}^{10}] = [1,0,...,0]$, then the loss becomes (7). According to *Inference 1*, the extremums of the loss function satisfy the following conditions:

$$\nabla_{W_{1+(l+k*\frac{t}{N}-1)\cdot w:(l+k*\frac{t}{N})\cdot w,o}} L = 0 \tag{12}$$

$$\nabla_b L = 0 \tag{13}$$

$$\nabla_{K^{l+k*\frac{t}{N}}} L = 0 \tag{14}$$

Therefore, we can get

$$I^{l+k*\frac{t}{N}} = 0 \tag{15}$$

$$-W_{1+(l+k*\frac{t}{N}-1)\cdot w:(l+k*\frac{t}{N})\cdot w,1} + \frac{1}{10}\left[\sum_{i=1}^{10} W_{1+(l+k*\frac{t}{N}-1)\cdot w:(l+k*\frac{t}{N})\cdot w,i}\right] = 0 \tag{16}$$

The value of *b* will not impact the conditions (before the nuclear transformation, i.e., (15) and (16)) of the loss function extremums, because we simplify the problem to the model with only one convolution layer and FC layer.

After the nuclear transformation, the model satisfies (1), (3), (4), (5) and (6). Suppose $y_{label} = [y_{label}^1,...,y_{label}^{10}] = [1,0,...,0]$, then the loss becomes (7). According to *Inference 1*, the extremums of the loss function satisfy the conditions (12)-(14).

Therefore, we get

$$\sum_{r=0}^{N-1} \beta_r I^{l+r*\frac{t}{N}} = 0 \tag{17}$$

$$\sum_{r=0}^{N-1} \left\{-W_{1+(l+r*\frac{t}{N}-1)\cdot w:(l+r*\frac{t}{N})\cdot w,1} + \frac{1}{10}\left[\sum_{i=1}^{10} W_{1+(l+k*\frac{t}{N}-1)\cdot w:(l+k*\frac{t}{N})\cdot w,i}\right]\right\} = 0 \tag{18}$$

When the loss *L* (before nuclear transformation) satisfies (15) and (16), $L = \ln(10)$; and when the loss $L'$ (after nuclear transformation) satisfies (17) and (18), $L = L' = \ln(10)$.

Intuitively speaking, (15) and (16) are more specific to the requirements of the conditions satisfied by the extreme values, however, (17) and (18) constructs a scope that includes the extremums that satisfy the conditions (15) and (16). In other words, the latter builds up more extremum points. Next, we will prove this by giving an example (based on *Theorem 2*).

In the variable space (where we fix variables $I^1,...,I^t$ and $W_{1:w,1},...,W_{1+(t-1)\cdot w:t\cdot w,1}$ of *V* to 0), we can transform (15) and (16) into homogeneous linear equations

$$B_{t\times 9t} V_{9t\times 1} = 0 \tag{19}$$

meanwhile, (17) and (18) can be transformed into equations

$$B'_{\frac{t}{N}\times 9t} V_{9t\times 1} = 0 \tag{20}$$

where $V_{11t\times 1} = \left[I^1,...,I^t,W_{1:w,1},...,W_{1+(t-1)\cdot w:t\cdot w,1},...,W_{1+(t-1)\cdot w:t\cdot w,10}\right]^T$.

All the elements in the solution set $V^1$ of (19) are the global minimums of *L*, and all the elements in the solution set $V^2$ of (20) are the global minimums of $L'$, under the condition of the solution set (where partial variables are fixed). According to *Theorem 2*, $r(B) < r(B')$ (*r* denotes the rank of the matrix), therefore, the solution number of global minimums (after the nuclear transformation) is more than that before. In a word, after nuclear transformation, the process of gradient descent would have faster convergence speed and higher efficiency, and more likely to get a more optimized solution, especially in the case with finite steps.

*C. Effect on FMs*

Fig.2 (a) and (b) are FMs obtained without using NS. By the weighted sum of multiple FMs containing noises, similar parts can be magnified, and the different parts can be weakened, thereby effectively reducing the noise in images of FMs. FMs in Fig.2 (c) and (d) are obtained using FNS with $N = 4$. It seems that the FM sharing is more like a voting system, from Fig.2 (d), which weakens the features of the few votes. Fig.2 (c) shows that by combining several abstract opinions of the image to be predicted (which is hard to recognize), we could produce better image views.

The comparison of Fig.3 (a) and (b) demonstrates that this method could improve the qualities of FMs learned by models notably. That is, with the help of NS, the convolution layer could obtain better features (with less noise).

IV. EXPERIMENT

*A. Experiement Setups*

We trained six neural networks with different depths for classification problems on datasets MNIST[1] [23] and CIFAR-10[2] [24] based on Tensorflow [25].

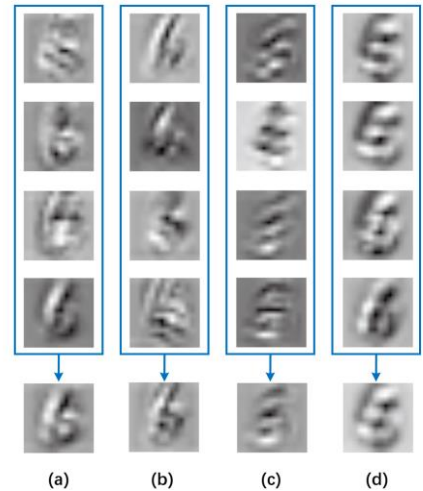

Fig. 2. Real cases of weighted sums of feature maps (weighted coefficient $\beta_r = 0.1, r = 0,..., N-1$). (a-b) Without using NS. (c-d) Using FNS.

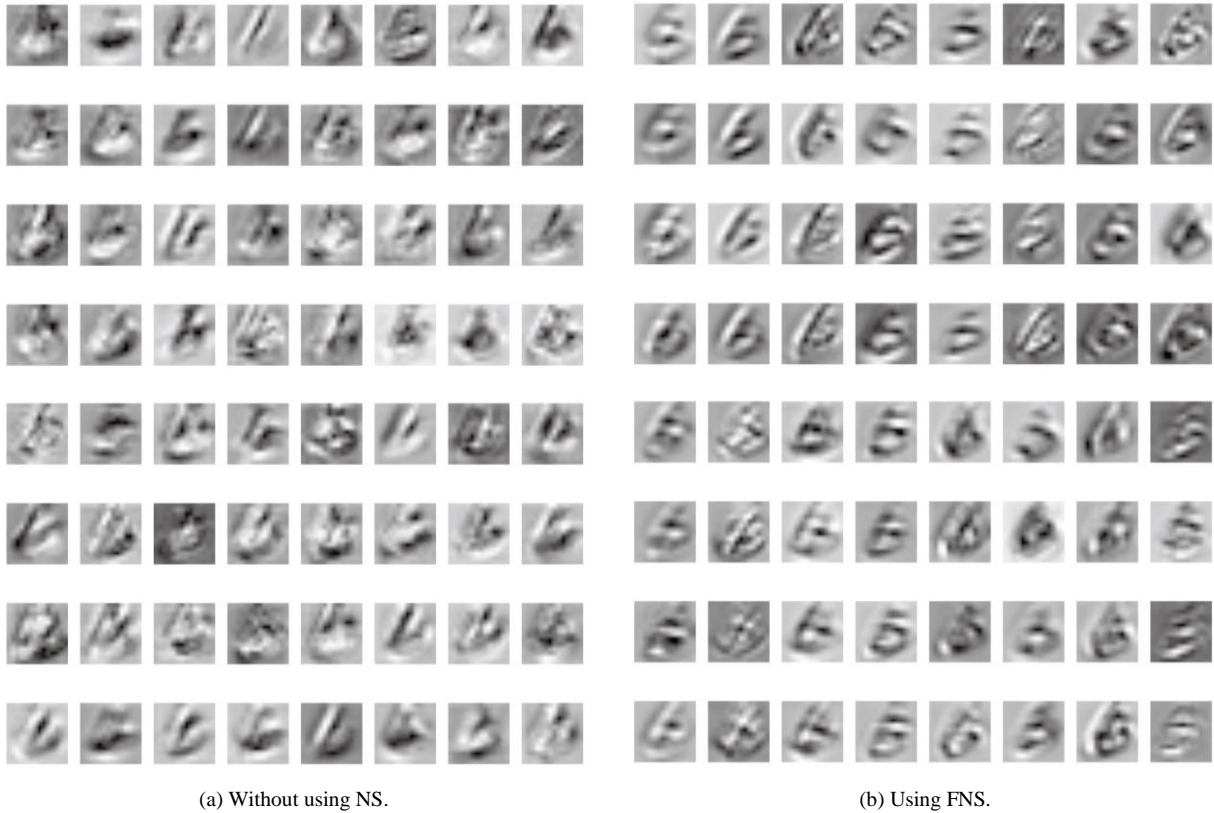

(a) Without using NS.  (b) Using FNS.

Fig. 3. FMs learned on MNIST by the second convolution layer of a network (the model has only two convolutional layers and uses the FC layer as the classification layer).

In the first instance, we set up a simple Multi-layer Perceptron (named SimpleMLP) and a simple Convolutional Neural Network (named SimpleCNN) to compare the NS with three fairly-known regularization methods on MNIST. The SimpleMLP has only one hidden layer of 784 nodes, and the SimpleCNN contains one convolution layer (the kernel size is $5\times5\times64$) and an FC layer (with one hidden layer of 1024 nodes). We use the minibatch size of 100 and use Rectified Linear Unit (ReLU) [26] as the activation function for the two simple networks. We divided the hidden nodes of SimpleMLP into four $14\times14$ FMs to use LRN on it. When using Dropout, we add it before FC layer, and the keep-rate is 0.5. We held out 10,000 random training images for validation, and the set of validation was then combined with the training set. Hyper-parameters were tuned on the validation set.

We then experimented on MNIST and CIFAR-10 datasets to observe the improvement in the performance of NS on neural networks at different depths. The LeNet-5 we used adds ReLU as the activation function and uses the minibatch size of 100, and adds Dropout before the FC layer. All networks used softmax loss function and trained 100 epochs on the entire training dataset. We use gradient descent optimization for SimpleMLP and use Adam [27] for the other networks. The other configuration of LeNet-5 and VGGNet [28] follows original settings. The weighting coefficients of NS are initialized to the same value. Every result is given with error bars by repeated the whole test procedure five times.

### B. Results and Discussion

Table I gives the comparison results of NS and other related regularization methods on MNIST. The results show that NS could notably improve the performance of SimpleMLP and SimpleCNN, compared with the other methods.

However, when coupling the convolution layer with the FC layer, improvements of performance is not as well as Dropout.

TABLE I. COMPARISON OF DIFFERENT REGULARIZATION METHODS ON MNIST (THE $\beta$ OF NS IS INITIALIZED TO 1 AND 0.25 FOR SIMPLEMLP AND SIMPLECNN, RESPECTIVELY).

| Method | | SimpleMLP | SimpleCNN |
|---|---|---|---|
| 2-fold | FNS | 95.84±0.07 | 99.14±0.03 |
| | TNS | **96.49±0.04** | 99.16±0.03 |
| 4-fold | FNS | 96.30±0.17 | 99.16±0.05 |
| | TNS | 96.46±0.06 | 99.17±0.06 |
| $L_2$ norm | | 95.04±0.09 | 99.10±0.04 |
| LRN | | 95.04±0.06 | 99.14±0.05 |
| Dropout | | 95.36±0.05 | **99.26±0.04** |
| Baseline | | 95.03±0.09 | 99.12±0.04 |

---

[1] http://yann.lecun.com/exdb/mnist/
[2] http://www.cs.toronto.edu/~kriz/cifar.html

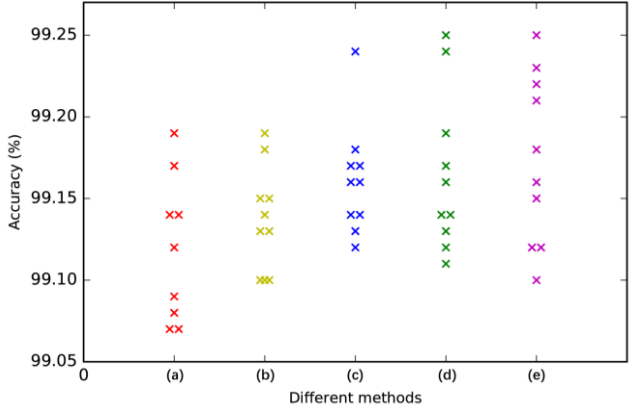

Fig. 4. Test accuracy distribution of SimpleCNN on MNIST (without using Dropout). (a) Baseline. (b-c) 2-fold FNS and TNS, respectively. (d-e) 4-fold FNS and TNS, respectively.

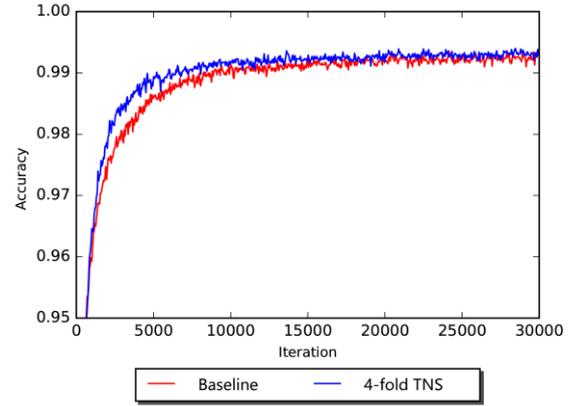

Fig. 5. Test accuracy curves of SimpleCNN (using Dropout) during training on MNIST.

TABLE II. RESULTS ON THE MNIST DATASET ($\beta$ IS INITIALIZED TO 1.00, 0.25, 0.25 AND 0.10 FOR SIMPLEMLP, SIMPLECNN, LENET-5 WHEN $N = 2$ AND LENET-5 WHEN $N = 4$, RESPECTIVELY).

| Method | | SimpleMLP | SimpleCNN | LeNet-5 |
|---|---|---|---|---|
| 2-fold | FNS | 95.84±0.07 | 99.29±0.02 | 99.43±0.04 |
| | TNS | **96.49±0.04** | 99.30±0.02 | 99.43±0.02 |
| 4-fold | FNS | 96.30±0.17 | 99.32±0.03 | 99.44±0.03 |
| | TNS | 96.46±0.06 | **99.32±0.02** | **99.45±0.03** |
| Baseline | | 95.03±0.09 | 99.26±0.04 | 99.39±0.03 |

TABLE III. RESULTS ON THE CIFAR-10 DATASET ($\beta$ IS INITIALIZED TO 0.02).

| Method | | VGG-11 | VGG-16 | VGG-19 |
|---|---|---|---|---|
| 2-fold | FNS | 74.59±0.04 | 75.99±0.24 | 76.10±0.53 |
| | TNS | 74.62±0.26 | 75.67±0.45 | 75.95±0.62 |
| 4-fold | FNS | **74.65±0.36** | **76.05±0.31** | 76.23±0.42 |
| | TNS | 74.21±0.49 | 75.65±0.54 | **76.36±0.47** |
| Baseline | | 73.73±0.34 | 75.59±0.64 | 75.96±0.35 |

The results of SimpleCNN and LeNet-5 (which use NS before Dropout) in Table II shows that NS could corporate with Dropout and have steady performance improvements.

To avoid the influence of random factors, we repeated the experiment of SimpleCNN (without using Dropout) 10 times to get the test accuracy distribution and plot it as Fig. 4. We can see that under the same number of experiments, NS could make the model perform more effective to achieve better results, and it seems that setting the weighted coefficient trainable and increase the value of $N$ could achieve better promotion effect (see Fig. 4). Fig. 5 illustrates that NS could speed up the convergence process of neural networks. Results in Table III show that NS could also achieve improved performance in deeper CNNs. From the overall results in this section, the trainable weighted coefficient $\beta$ and a larger $N$ could help pursuing better performance. The initialization of $\beta$ is very important when using NS. With the deepening in the depth of network, the initialization value of $\beta$ should be smaller.

## V. CONCLUSION

In this paper, we introduce a simple method named N-fold Superposition to the CNNs. Through the theoretical analysis, we have elaborately illustrated how NS improves the dependency of FM and hidden weights to improve the coupling between the convolution layer and the FC layer. Moreover, we proved this method could construct more global minima values to make it easier for networks to converge to the optimal point, and experiments prove the effectiveness of NS in reducing the noise in FMs, speeding up the convergence process and improving the performance.

ACKNOWLEDGMENT

This work is supported by the National Natural Science Foundation of China (NSFC No.61601513).

REFERENCES

[1] Lee, Han S., Heechul Jung, Alex A. Agarwal, and Junmo Kim. "Can Deep Neural Networks Match the Related Objects?: A Survey on ImageNet-trained Classification Models." arXiv preprint arXiv:1709.03806 (2017).

[2] Liang, Chang, Deng Xiaoming, and Zhou Mingquan. "Convolutional neural networks in image understanding." Acta Automatica Sinica 42, no. 9 (2016): 1300-1312.

[3] Murugan, Pushparaja, and Shanmugasundaram Durairaj. "Regularization and optimization strategies in deep convolutional neural network." arXiv preprint arXiv:1712.04711(2017).

[4] Lin, Min, Qiang Chen, and Shuicheng Yan. "Network in network." arXiv preprint arXiv:1312.4400 (2013).

[5] Szegedy, C., W. Liu, Y. Jia, P. Sermanet, S. Reed, and D. Anguelov. "& Rabinovich, A. Going deeper with convolutions." In Proceedings of the IEEE conference on computer vision and pattern recognition, pp. 1-9. 2015.

[6] Szegedy, Christian, Sergey Ioffe, Vincent Vanhoucke, and Alexander A. Alemi. "Inception-v4, inception-resnet and the impact of residual connections on learning." In AAAI, vol. 4, p. 12. 2017.

[7] Boureau, Y-Lan, Jean Ponce, and Yann LeCun. "A theoretical analysis of feature pooling in visual recognition." In Proceedings of the 27th

international conference on machine learning (ICML-10), pp. 111-118. 2010.

[8] He, Kaiming, Xiangyu Zhang, Shaoqing Ren, and Jian Sun. "Deep residual learning for image recognition." In Proceedings of the IEEE conference on computer vision and pattern recognition, pp. 770-778. 2016.

[9] Szegedy, Christian, Vincent Vanhoucke, Sergey Ioffe, Jon Shlens, and Zbigniew Wojna. "Rethinking the inception architecture for computer vision." In Proceedings of the IEEE Conference on Computer Vision and Pattern Recognition, pp. 2818-2826. 2016.

[10] Hu, Jie, Li Shen, and Gang Sun. "Squeeze-and-excitation networks." arXiv preprint arXiv:1709.01507 (2017).

[11] Ranjan, Rajeev, Carlos D. Castillo, and Rama Chellappa. "L2-constrained softmax loss for discriminative face verification." arXiv preprint arXiv:1703.09507 (2017).

[12] Liu, Weiyang, Yandong Wen, Zhiding Yu, and Meng Yang. "Large-Margin Softmax Loss for Convolutional Neural Networks." In ICML, pp. 507-516. 2016.

[13] Krogh, Anders, and John A. Hertz. "A simple weight decay can improve generalization." In Advances in neural information processing systems, pp. 950-957. 1992.

[14] Hinton, Geoffrey E., et al."Improving neural networks by preventing co-adaptation of feature detectors." arXiv preprint arXiv:1207.0580 (2012).

[15] Srivastava, Nitish, Geoffrey Hinton, Alex Krizhevsky, Ilya Sutskever, and Ruslan Salakhutdinov. "Dropout: A simple way to prevent neural networks from overfitting." The Journal of Machine Learning Research 15, no. 1 (2014): 1929-1958.

[16] Ioffe, Sergey, and Christian Szegedy. "Batch normalization: Accelerating deep network training by reducing internal covariate shift." arXiv preprint arXiv:1502.03167 (2015).

[17] Glorot, Xavier, and Yoshua Bengio. "Understanding the difficulty of training deep feedforward neural networks." In Proceedings of the thirteenth international conference on artificial intelligence and statistics, pp. 249-256. 2010.

[18] He, Kaiming, Xiangyu Zhang, Shaoqing Ren, and Jian Sun. "Delving deep into rectifiers: Surpassing human-level performance on imagenet classification." In Proceedings of the IEEE international conference on computer vision, pp. 1026-1034. 2015.

[19] Krizhevsky, Alex, Ilya Sutskever, and Geoffrey E. Hinton. "Imagenet classification with deep convolutional neural networks." In Advances in neural information processing systems, pp. 1097-1105. 2012.

[20] Zhang, Chaoyun, Pan Zhou, Chenghua Li, and Lijun Liu. "A convolutional neural network for leaves recognition using data augmentation." In Computer and Information Technology; Ubiquitous Computing and Communications; Dependable, Autonomic and Secure Computing; Pervasive Intelligence and Computing (CIT/IUCC/DASC/PICOM), 2015 IEEE International Conference on Computer and Information Technology, pp. 2143-2150. IEEE, 2015.

[21] Neyshabur, Behnam, Ryota Tomioka, and Nathan Srebro. "Norm-based capacity control in neural networks." In Conference on Learning Theory, pp. 1376-1401. 2015.

[22] Prechelt, Lutz. "Automatic early stopping using cross validation: quantifying the criteria." Neural Networks 11, no. 4 (1998): 761-767.

[23] LeCun, Yann, Léon Bottou, Yoshua Bengio, and Patrick Haffner. "Gradient-based learning applied to document recognition." Proceedings of the IEEE 86, no. 11 (1998): 2278-2324.

[24] Alex Krizhevsky and Geoffrey Hinton. Learning multiple layers of features from tiny images. Technical report, Technical report, University of Toronto, 2009.

[25] Abadi, Martín, Paul Barham, Jianmin Chen, Zhifeng Chen, Andy Davis, Jeffrey Dean, Matthieu Devin et al. "TensorFlow: A System for Large-Scale Machine Learning." In OSDI, vol. 16, pp. 265-283. 2016.

[26] Nair, Vinod, and Geoffrey E. Hinton. "Rectified linear units improve restricted boltzmann machines." In Proceedings of the 27th international conference on machine learning (ICML-10), pp. 807-814. 2010.

[27] Kingma, Diederik P., and Jimmy Ba. "Adam: A method for stochastic optimization." arXiv preprint arXiv:1412.6980(2014).

[28] Simonyan, Karen, and Andrew Zisserman. "Very deep convolutional networks for large-scale image recognition." arXiv preprint arXiv:1409.1556 (2014).